\documentclass{ecai2014}
\usepackage{times}
\usepackage{graphicx}
\usepackage{latexsym}
\usepackage{amsmath,amsthm,amssymb,amsbsy}
\usepackage{url}
\usepackage[mathscr]{eucal}
\newcommand{\bs}[1]{\boldsymbol{#1}}
\newcommand{\tensor}[1]{\bs{\mathscr{\MakeUppercase{#1}}}} 
\newcommand{\tA}{\tensor{A}}

\DeclareGraphicsExtensions{.jpg,.png,.pdf}
\begin{document}

\title{On tensor rank of conditional probability tables\\ in Bayesian networks}

\author{Ji\v{r}\'{\i} Vomlel\institute{Institute of Information Theory and Automation of the AS CR, Prague, Czech Republic, email: vomlel@utia.cas.cz} \and Petr Tichavsk\'y \institute{Institute of Information Theory and Automation of the AS CR, Prague, Czech Republic, email: tichavsk@utia.cas.cz}
}

\maketitle
\bibliographystyle{ecai2014}

\begin{abstract}
A difficult task in modeling with Bayesian networks is the elicitation of numerical
parameters of Bayesian networks. A large number of parameters is needed to specify
a conditional probability table (CPT) that has a larger parent set. 
In this paper we show that, most CPTs from real applications of Bayesian networks
can actually be very well approximated by tables that require substantially less parameters. 
This observation has practical consequence not only for model elicitation but also
for efficient probabilistic reasoning with these networks.
\end{abstract}

\section{INTRODUCTION}

On the most cumbersome tasks in modeling with Bayesian networks 
is the elicitation of conditional probabilities. These probabilities
can be estimated from collected data or elicited with the help of domain
experts. A general conditional probability table (CPT), $P(X_i | pa(X_i))$,
where $X_i$ is a discrete random variable and $pa(X_i)$ are its parent variables,
requires $n_i = (|X_i|-1) \cdot \prod_{X_j \in pa(X_i)} |X_j|$ parameters.
The cardinality $|X|$ of a variable $X$ is the number of values the variable $X$ can take.
Since $n_i$ is exponential with respect to number of parents the task of probability elicitation
can be quite demanding for tables with a large parent set.

To ease the task of probability elicitation special types of CPTs were proposed.
They include, so called, canonical models~\cite{diez-druzdzel-2006} whose typical
examples are noisy-or, noisy-max and noisy-threshold, etc.
For these special types of CPTs the number of parameters is often linear or quadratic
with respect to the number of parents. Furthermore, it is possible to perform more
efficient computations with them~\cite{olesen-1989, diez-2003, savicky-2007, vomlel-2013}.

On the other hand the vast majority of Bayesian networks used in real applications
that are available in Bayesian network repositories uses only general CPTs.
The main motivation of this paper is to show that actually many of these general
tables have a hidden simple structure that requires a lower number of parameters
than general CPTs. We will apply the CP-tensor decomposition~\cite{harshman-1970, carrol-1970}
for CPTs~\cite{savicky-2007, vomlel-2013} of models from a BN repository. We will see
that many of the CPTS of these Bayesian networks can be approximated pretty well
by tables with a low rank. 

This observation has two practical consequences.
First, more attention should be paid to discovering the internal structure of CPTs 
that are either estimated from data or elicited with the help of domain experts.
Second, the internal structure of CPTs can be exploited for more efficient
probabilistic inference with these networks.

The rest of this paper is organized as follows.
In Section~\ref{sec-cp-decomp} we briefly introduce the CP tensor decomposition.
The results of experiments with the state-of-the-art numerical algorithms for 
the CP-tensor decomposition are reported in Section~\ref{sec-experiments}.
We conclude the paper by a summary in Section~\ref{sec-conclusions}.

\section{CP TENSOR DECOMPOSITION}\label{sec-cp-decomp}

Each CPT can be viewed as a tensor. 
A tensor is simply a mapping $\tA: \mathbb{I} \rightarrow {\mathbb R}$, 
where $\mathbb{I} = I_1 \times \ldots \times I_k$, $k$ is a natural number 
called the order of tensor $\tA$, and $I_j, j=1,\ldots,k$ are index sets. 
Typically, $I_j$ are sets of integers of cardinality $n_j$. 
Then we can say that tensor $\tA$ has dimensions $n_1, \ldots, n_k$.
Tensor $\tA$ has rank one if it can be written as an outer product
of vectors:
\begin{eqnarray*}
\tA & = & \bs{a}_1 \otimes \ldots \otimes \bs{a}_k 
\end{eqnarray*}
with the outer product $\otimes$ 
being defined for all $(i_1,\ldots,i_k) \in I_1 \times \ldots \times I_k$ as
$\tA_{i_1,\ldots,i_k} = a_{1,i_1} \cdot \ldots \cdot a_{k,i_k}$,
where $\bs{a}_j=\left(a_{j,i}\right)_{i \in I_j}$,  $j=1,\ldots,k$  are real valued vectors.

CP-tensor decomposition~\cite{harshman-1970, carrol-1970} is 
decomposition of a tensor into a sum of $r$ tensors of rank one.
The minimum number $r$ such that the decomposition is possible is called the rank
of tensor $\tA$.

In~\cite{savicky-2007} the CP tensor decomposition (called tensor rank-one decomposition there) 
was applied to CPTs of canonical models. See~\cite{savicky-2007}~or~\cite{vomlel-2013} 
for a more detailed explanation, which is out of scope of this paper.
Instead, we will just illustrate savings one may get with the CP tensor decomposition of CPTs. 
Assume a CPT defined for a binary variable with $k-1$ binary parents.
To fully specify such a table we need $2^{k-1}$ parameters. If this table can be 
approximated by a table with rank $r$ we need $k(r-1)+r$ parameters instead.
If $r=2$ then we need $k+2$ parameters. We can immediately see that instead of 
an exponential number of parameters we need just a linear number of them.

\section{NUMERICAL EXPERIMENTS}\label{sec-experiments}

For the experiments we used 15 Bayesian networks from a Bayesian network repository\footnote{We
used networks from the repository at \url{http://www.bnlearn.com/bnrepository/}, where networks 
from other repositories after a quality-check and fixing are stored.}, namely: 

\emph{alarm.net, hailfinder.net, barley2.net,
pathfinder.net, munin1.net, munin2.net, munin3.net,
munin4.net, mildew.net, hepar2.net, andes.net, win95pts.net,
water.net, link.net, and insurance.net}.

In the experiments we considered all CPTs for variables having at least three parents.
It was 492 CPTs in total. For the CP tensor decomposition we used the
Fast Damped Gauss-Newton (Levenberg-Marquard) algorithm~\cite{phan-2013} 
implemented in the Matlab package TENSORBOX~\footnote{Tensorbox is freely available at \url{http://www.bsp.brain.riken.jp/~phan/tensorbox.php}.}. In the first experiment we used
this algorithm with 100 different random starting points and one precomputed (using nvec method).
In the second experiment we used 10+1 starting points.

\begin{figure}[htb]
\centerline{\includegraphics[width=0.65\columnwidth]{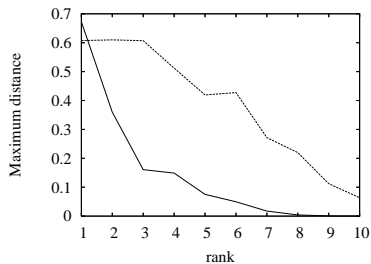}}
\caption{Maximum distance of the approximation to the original CPT as a function of 
its rank.} \label{fig-3x3x4x3}
\end{figure}

First, we present the results for an example of a $3 \times 3 \times 4 \times 3$ CPT from the \emph{hailfinder.net} network. In this CPT the child 
variable \emph{Boundaries} (3 states) has three parents 
\emph{OutflowFrMt} (3 states), \emph{MorningBound} (3 states), and \emph{WndHodograph} (4 states).
In Figure~\ref{fig-3x3x4x3} we compare the maximum distance 
$$\max_{(i_1,\ldots,i_k)} | \tensor{\hat{A}}_{i_1,\ldots,i_k} - \tA_{i_1,\ldots,i_k}|$$
of the approximation $\tensor{\hat{A}}$ to the original CPT $\tA$ as a function of the rank of the approximation.
We can see that the distance rapidly decreases with rank (full line), 
which is not the case of a table of the same dimensions with random values
(dashed line).

\begin{figure}[htb]
\centerline{\includegraphics[width=0.65\columnwidth]{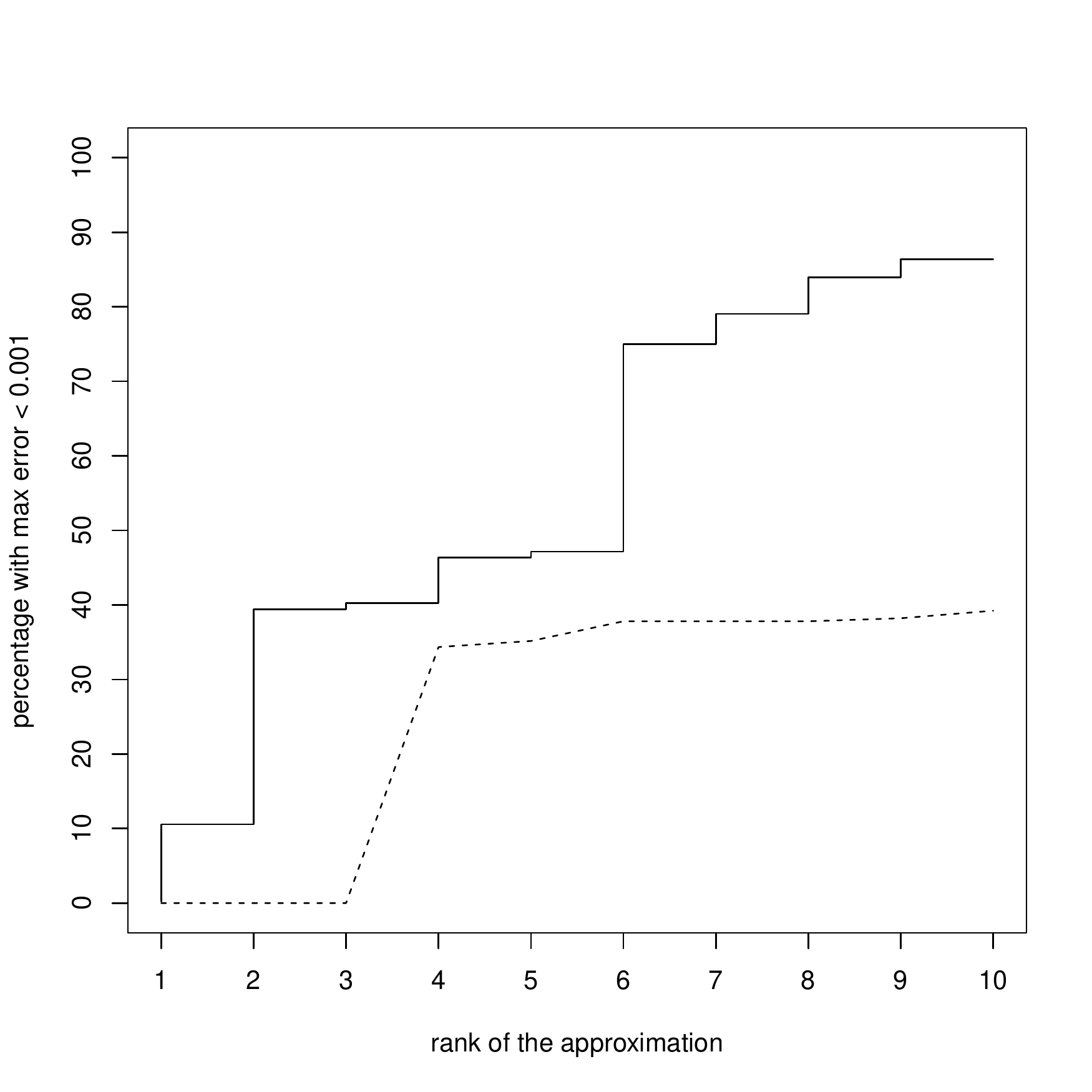}}\
\caption{Percentage of CPTs that can be approximated with a maximum error smaller than 0.001 as a function
of the rank of the approximation.} \label{fig-percentage}
\end{figure}

Next, we summarize results for all $492$ tables from $15$ different Bayesian networks.
In Figure~\ref{fig-percentage} we present the plot describing dependence of 
percentage of CPTs that can be approximated by tables of a given rank 
with maximum error (taken over all possible configurations of parents)
smaller than $0.001$. See the full line.
From this figure we can see that about $40\%$ of the CPTs can be pretty well
approximated by tables of rank 2, about $75\%$ by tables of rank 6, and only about $15\%$ require
rank more than 10 to get a very good fit.

One may conjecture that any CPT can be approximated well by a table with a similarly low rank as
CPTs of networks from repositories.
To show that this is not the case we repeated the computations for CPTs of the same dimensions but with their entries replaced by random numbers. See dashed line in Figure~\ref{fig-percentage}. 
We can see that it is not possible to approximate randomly created CPTs with similarly low rank
as for CPTs of Bayesian networks from the repository.

\section{CONCLUSIONS}\label{sec-conclusions}

We presented numerical experiments on $15$ Bayesian network models from real applications.
The results suggest that most of the conditional probability tables can be very well
approximated by tables that have lower rank than one would expect from a general table of the same dimensions.
Our results are in line with results presented in~\cite{zagorecki-2013} where the authors
experimentally verified that noisy-max provides a surprisingly good fit for as many as
$50\%$ of CPTs in two of the three Bayesian networks they tested. 

The low rank approximation should be exploited not only in the model elicitation by using 
a compact parametrization of the internal structure of a CPT but also during the probabilistic inference.
We conjecture that some low rank approximations of CPTs may actually correspond better  
to what a domain expert intended to model in the constructed CPT.
One reason might be that when filling numerical values into a CPT with more than two parents it 
is quite demanding for the model creator to provide exact numerical values. 
Another reason might be that numerical values in the CPTs are often provided with
a lower numerical precision, typically with two decimal points.

\ack This work was supported by the Czech Science Foundation through projects 
No. 13--20012S  and No. 14--13713S.

\bibliography{ecai2014}
\end{document}